\documentclass[11pt]{article}
\usepackage{amsmath,amssymb,amscd,amsthm}
\usepackage{graphicx}
\usepackage{dcolumn}
\usepackage{bm}
\usepackage{ifthen}
\usepackage{rays_defs_18}

\usepackage[T1]{fontenc}

\usepackage[utf8]{inputenc} 
\usepackage[T1]{fontenc}    
\usepackage{url}            
\usepackage{booktabs}       
\usepackage{amsfonts}       
\usepackage{nicefrac}       
\usepackage{microtype}      
\usepackage{xcolor}         

\usepackage{hyperref}
\hypersetup{
    bookmarks=true,         
    unicode=false,          
    pdftoolbar=true,        
    pdfmenubar=true,        
    pdffitwindow=false,     
    pdfstartview={FitH},    
    pdfauthor={Fatih Furkan YILMAZ},     
    pdfsubject={Subject},   
    pdfcreator={Fatih Furkan YILMAZ},   
    pdfproducer={Producer}, 
    pdfnewwindow=true,      
    colorlinks=true,       
    linkcolor=red,          
    citecolor=green,        
    filecolor=magenta,      
    urlcolor=cyan           
}

\usepackage{filecontents}

\usepackage[
backend=bibtex,
url=false,isbn=false,doi=false,
date=year,
hyperref=auto,
style=alphabetic,
natbib=true,
sorting=nty,
firstinits=true,
maxnames=2, maxbibnames=10,
]{biblatex}

\bibliography{./bibliography.bib} 

\usepackage{bbm}

\usepackage{comment}

\usepackage{tikz}
\usepackage{pgfplots}
\usepgfplotslibrary{groupplots} 
\usetikzlibrary{pgfplots.groupplots}
\usetikzlibrary{matrix,arrows,decorations.pathmorphing}
\usepackage{pgfplots,pgfplotstable}
\usepgfplotslibrary{fillbetween}
\pgfplotsset{compat=1.14}

\usetikzlibrary{matrix,arrows,decorations.pathmorphing}

\usepackage{color,framed,bm}

\usepackage{colortbl}
\definecolor{tblblue}{RGB}{101,124,191}
\definecolor{tblred}{rgb}{1,0.93,0.93}
\definecolor{DarkBlue}{rgb}{0,0,0.7} 
\definecolor{BrickRed}{RGB}{203,65,84}
\definecolor{mygray}{gray}{0.6}

\newtheorem{lemma}{Lemma}

\newtheorem{theorem}{Theorem}

\newtheorem{proposition}{Proposition}

\newcommand\iter{t}

\newcommand\risk{R}

\newcommand\vtheta{{\bm \theta}}

\newcommand\mSigma{\bm \Sigma}

\newcommand\mLambda{\bm \Lambda}

\newcommand\setD{\mathcal D}

\newcommand\mystackrel[2]{\stackrel{\text{#1}}{#2}}

\newcommand\relu{\mathrm{relu}}

\long\def\comment#1{}

\pgfplotsset{select coords between index/.style 2 args={
    x filter/.code={
        \ifnum\coordindex<#1\fi
        \ifnum\coordindex>#2\fi
    }
}}

\begin{document}

\begin{center}

{\bf{\LARGE{
    Regularization-wise double descent: Why it occurs and how to eliminate it
}}}

\vspace*{.2in}

{\large{
    \begin{tabular}{cccc}
        Fatih Furkan Yilmaz$^{\ast}$ 
        and 
        Reinhard Heckel$^{\ast,\dagger}$
    \end{tabular}
}}

\vspace*{.05in}

\begin{tabular}{c}
    $^\ast$Dept. of Electrical and Computer Engineering, Rice University\\
    $^\dagger$Dept. of Electrical and Computer Engineering, Technical University of Munich
\end{tabular}


\vspace*{.1in}

\end{center}


\begin{abstract}
The risk of overparameterized models, in particular deep neural networks, is often double-descent shaped as a function of the model size. Recently, it was shown that the risk as a function of the early-stopping time can also be double-descent shaped, and this behavior can be explained as a super-position of bias-variance tradeoffs. 
In this paper, we show that the risk of explicit L2-regularized models can exhibit double descent behavior as a function of the regularization strength, both in theory and practice. We find that for linear regression, a double descent shaped risk is caused by a superposition of bias-variance tradeoffs corresponding to different parts of the model and can be mitigated by scaling the regularization strength of each part appropriately. 
Motivated by this result, we study a two-layer neural network and show that double descent can be eliminated by adjusting the regularization strengths for the first and second layer. Lastly, we study a 5-layer CNN and ResNet-18 trained on CIFAR-10 with label noise, and CIFAR-100 without label noise, and demonstrate that all exhibit double descent behavior as a function of the regularization strength.
\end{abstract}

\section{Introduction}
\label{sec:intro}

The bias-variance tradeoff has long been a useful principle for selecting and tuning machine learning models. This principle suggests to choose a model sufficiently large to have low bias, but not too large to have small variance. 
In practice, however, machine learning models seemingly operate beyond this tradeoff. Deep neural networks operate in the overparameterized regime where the model is capable of expressing any given signal, even random noise~\citep{zhang_UnderstandingDeepLearning_2016}, but still generalize well. 
Increasing the model size beyond the interpolation point often decreases the test error beyond the classical U-shaped curve, hence forming a double descent shaped risk curve~\citep{opper_StatisticalMechanicsLearning_1995,belkin2019reconciling}.


Machine learning algorithms are often regularized during training to improve performance, and similar to model size, the amount of regularization can control a bias-variance tradeoff. Indeed, recently, double descent behavior was reported as a function of training epochs and weight decay~\citep{nakkiran_DeepDoubleDescent_2020}.  
Understanding such double descent behavior is important because it can be critical for good performance, especially for learning from noisy labels~\citep{arpit_CloserLookMemorization_2017,yilmaz_ImageRecognitionRaw_2020}.

The perhaps most popular regularization technique is to add an explicit $\ell_2$-norm penalty to the training loss (i.e., a term $\lambda \norm[2]{\vtheta}^2$), or training with weight decay, in deep learning semantics. Double descent as a function of the regularization parameter $\lambda$ has been reported for a ResNet-18 network trained on CIFAR-10 with label noise~\citep[Figure 22]{nakkiran_DeepDoubleDescent_2020}, but a theoretical understanding and a more extensive empirical study covering a variety of models is still lacking.

\begin{figure}[t]
    \begin{center}
    \begin{tikzpicture}

    \begin{groupplot}[
    width=7cm,
    height=5cm,
    group style={
        group name=mcnn_regularization,
        group size=3 by 1,
        xlabels at=edge bottom,
        ylabels at=edge left,
        yticklabels at=edge left,
        xticklabels at=edge bottom,
        horizontal sep=0.5cm, 
        vertical sep=1.8cm,
        },
    legend cell align={left},
    legend columns=1,
    legend style={
                at={(1,1)}, 
                anchor=south east,
                draw=black!50,
                fill=none,
                font=\small
            },
    /tikz/every even column/.append style={column sep=0.1cm},
    /tikz/every row/.append style={row sep=0.3cm},
    colormap name=viridis,
    cycle list={[colors of colormap={50,350,750}]},
    tick align=outside,
    tick pos=left,
    x grid style={white!69.01960784313725!black},
    xtick style={color=black},
    y grid style={white!69.01960784313725!black},
    ytick style={color=black},
    extra y ticks={0.2119}, 
    extra y tick style={%
        grid=major,
    },
    extra y tick labels=,
    xlabel={$1/\lambda$}, 
    ylabel={test error},
    ymin=0.1, ymax=0.5,
    xmode=log,
    every axis plot/.append style={thick},
    ]
    
    
    \nextgroupplot[]
    
    \addplot +[mark=none] table[x expr=1/\thisrow{lambda},y expr=1 - \thisrow{test}/100]{./fig/data/mcnn_lambda_error_extended.txt};

    \nextgroupplot[xlabel=epoch]
    
    \addplot +[mark=none] table[x=epoch,y expr=1 - \thisrow{test}/100]{./fig/data/mcnn_double_descent.txt};
    
    \end{groupplot}
    
    \end{tikzpicture}
    \end{center}
    \vspace{-0.2cm}
    \caption{\label{fig:mcnn_lambda}
    Test performance of a 5-layer convolution network when trained on the CIFAR-10 dataset with 20\% label noise.
    {\bf Left:} 
    Performance as a function of the inverse regularization parameter $1/\lambda$ if the network is trained with $\ell_2$-regularization until convergence.  
        {\bf Right:}
    Performance as a function of the training epochs when the network is trained without $\ell_2$ regularization.
    {\bf Both:} 
    In both cases, the network exhibits double descent behavior as a function of both the regularization by $\lambda$ and regularization by early stopping the training.
    }
    \vspace{-0.5cm}
\end{figure}
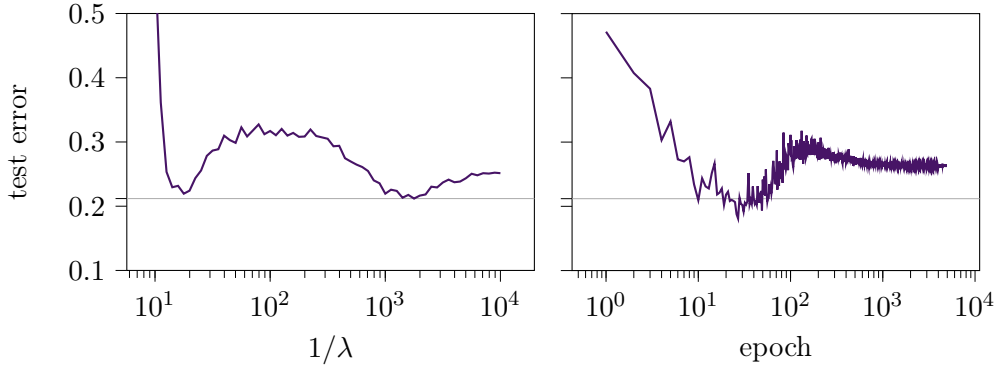

In this paper, we therefore study the risk of $\ell_2$-regularized models as a function of the regularization strength $\lambda$, both in theory and practice.
Our contributions are as follows:
\begin{itemize}
\item Our empirical results show that various neural networks regularized with an $\ell_2$-penalty $\lambda \norm[2]{\vtheta}^2$ can exhibit double descent shaped risk curve as a function of 
$1/\lambda$.  That is, the risk or test error first decreases, then increases, and then decreases again as a function of $1/\lambda$ (see Figure~\ref{fig:mcnn_lambda}). This frequently occurs when training on noisy data, but can also occur when training standard models (a CNN) on a standard noise-less dataset (CIFAR-100).

\item Next, we consider a linear ridge regression model and theoretically characterize the risk as a function $\lambda$. We show that when the features have different scales, similarly to early-stopped least squares studied in~\citep{heckel_EarlyStoppingDeep_2020}, the risk of the ridge regression solution as a function of $1/\lambda$ is a superposition of bias-variance tradeoffs, which yields a double descent behavior. 

\item Finally, we consider a non-linear two-layer neural network and provide numerical examples where double descent occurs as a function of $1/\lambda$ and, motivated by our theory in the linear case, eliminate the double descent by utilizing differently scaled $\lambda$ values for the two layers. Eliminating double descent is interesting as it typically improves the performance of the best model.
\end{itemize}

While conceptually our results for explicit $\ell_2$-regularization parallel those for early stopping developed in our earlier paper~\citep{heckel_EarlyStoppingDeep_2020}, early-stopping and $\ell_2$-regularization often behave quite differently: 
Figure~\ref{fig:mcnn_lambda} shows the test error of a 5-layer CNN as a function of $1/\lambda$ when trained on the noisy CIFAR-10 dataset, and contrasts this to the test error as a function of the training epochs (with no $\ell_2$-regularization). Note that the test error as a function of (inverse) regularization strength exhibits a double descent behavior and regularization with early stopping exhibits a double descent behavior (as shown before by~\citep{nakkiran_DeepDoubleDescent_2020}), but the effect of $\ell_2$-regularization and early stopping is not the same as the $\ell_2$-regularization allows attaining the same best-case performance in two distinct regimes, whereas the early stopped risk does not.



\section{Related works}
Double descent as a function of the model size has been theoretically established for linear regression~\citep{hastie_SurprisesHighDimensionalRidgeless_2019,belkin2020two,mitra2019UnderstandingOverfittingPeaks} and for random feature regression~\citep{mei_GeneralizationErrorRandom_2019, dascoli2020DoubleTroubleDouble}. 
Double descent has also been studied as a function of training time~\citep{heckel_EarlyStoppingDeep_2020,zhang2021optimization} and sample complexity~\citep{nakkiran2019MoreDataCan}.
\citet{nakkiran_DeepDoubleDescent_2020} have provided several empirical examples of epoch-wise, sample-wise, and regularization-wise double descent for deep networks.  
Beyond double descent, multiple decent has also been shown and characterized in the paper~\citep{liang2020MultipleDescentMinimumNorm,heckel_EarlyStoppingDeep_2020}.


A recent line of theoretical model-wise double descent works studied the behavior of the risk, specifically by decomposing the risk into bias and variance terms ~\citep{jacot_ImplicitRegularizationRandom_2020, yang_RethinkingBiasVarianceTradeoff_2020,dascoli2020DoubleTroubleDouble,liang2020JustInterpolateKernel,liang2020MultipleDescentMinimumNorm}.$\,$
Several works have further decomposed bias-variance terms with respect to the different sources of randomness in training, such as the optimization process or data distribution~\citep{neal_ModernTakeBiasVariance_2018, adlam2020UnderstandingDoubleDescent}.$\,$
Our model also relies on the interaction between the data and the model parameters to study double descent.

For epoch-wise double descent,~\citet{heckel_EarlyStoppingDeep_2020} characterized the risk as a function of the training time as a superposition of multiple bias-variance tradeoffs, which yields double descent for misaligned features. For a setup with misaligned features, we show an analogous result where we decompose the risk as a function of $1/\lambda$ as a superposition of bias-variance tradeoffs. 

Generalization and training dynamics of deep networks with $\ell_2$ regularization in the form of {\it weight decay} has been a topic of interest, particularly regarding finding optimal setups, such as finding the optimal weight matrix based on the data prior for weighted regularization~\citep{wu2020OptimalWeightedEll}. 
\citet{nakkiran_OptimalRegularizationCan_2020} have shown that optimal $\ell_2$ regularization can mitigate model-wise and sample-wise double descent, analytically for linear regression and empirically for CNNs.

Many works used neural-tangent-kernels (NTKs)~\citep{jacot_NeuralTangentKernel_2018}, 
to study the double descent behavior, as a function of the network width~\citep{adlam2020NeuralTangentKernel} and training epochs~\citep{heckel_EarlyStoppingDeep_2020}, as well as to understand the dynamics of $\ell_2$-regularized neural network training~\citep{wei2020RegularizationMattersGeneralization,lewkowycz2021TrainingDynamicsDeep}.
~\citet{lewkowycz2021TrainingDynamicsDeep} demonstrated that the NTK deviates significantly from initialization after a time that is inversely proportional to the regularization strength.


\section{Ridge regression risk as a function of the regularization parameter}
\label{sec:ridgeregression}

We start with studying the risk of the ridge regression estimator with regularization parameter $\lambda$, for fitting a linear model to data generated by a Gaussian linear model.  
We show that the risk as a function of $1/\lambda$ is a superposition of U-shaped bias-variance tradeoffs. If the features of the Gaussian linear model have different scales, those bias-variance tradeoff curves can add up to a double (or multiple) descent shaped risk curve.

\subsection{Data model and risk}
\label{sec:datamodel}

We consider the same linear regression setup as~\citet{heckel_EarlyStoppingDeep_2020}. 
Consider a regression problem, and suppose data is generated from a Gaussian linear model as
$
y = \innerprod{\vx}{\vtheta^\ast} + z,
$
where $\vx \in \reals^d$ is a zero-mean Gaussian feature vector with diagonal co-variance matrix $\mSigma = \diag(\sigma_1^2,\ldots,\sigma_d^2)$, and $z$ is independent, zero-mean Gaussian noise with variance $\sigma^2$.
We are given a training set $\setD = \{(\vx_1,y_1),\ldots,(\vx_n,y_n)\}$ consisting of $n$ data points drawn iid from this Gaussian linear model.

Consider a linear estimator parameterized by a vector $\hat \vtheta \in \reals^d$ which predicts the label associated with a feature vector $\vx$ as $\hat y = \innerprod{\vx}{\hat \vtheta}$.
The (mean-squared) risk of this estimator is
\begin{align*}
\risk(\hat \vtheta)
&=
\EX{\left(y - \innerprod{\vx}{\hat \vtheta}\right)^2},
\end{align*}
where the expectation is over an example $(\vx,y)$ drawn independently (of the training set) from the underlying linear model. 
The risk of the estimator can be written as a function of the variances of the features and of the coefficients of the underlying true linear model, $\vtheta^\ast = [\theta_1^\ast,\ldots, \theta_d^\ast]$, as
\begin{align}
\risk(\hat \vtheta)
&=
\sigma^2
+
\sum_{i=1}^d \sigma_i^2 (\theta^\ast_i - \hat \theta_i)^2.
\label{eq:expressioniterates}
\end{align}

\subsection{Risk of the ridge regression estimator}

Consider the ridge regression estimator defined as 
\begin{align*}
\hat \vtheta_{\lambda}
=
\arg \min_{\vtheta}
\frac{1}{2}
\sum_{i=1}^n (\innerprod{\vx_i}{\vtheta} - y_i)^2
+ \frac{\lambda}{2} \norm[2]{\vtheta}^2.
\end{align*}
We show that in the underparameterized regime, where $d \ll n$, the risk of the ridge regression estimate, $\risk(\hat \vtheta_\lambda)$, is very well approximated by
\begin{align}
    \label{eq:defbarR}
    \bar \risk(\tilde \vtheta_\lambda)
    =
    \sigma^2 + \sum_{i=1}^d
    \underbrace{
    \sigma_i^2 (\theta_i^\ast)^2 
    \left( \frac{\lambda}{\sigma_i^2 + \lambda} \right)^2  
    +
    \frac{\sigma^2}{n} 
    \sigma_i^2
    \left( \frac{\sigma_i}{\sigma_i^2 + \lambda} \right)^2
    }_{V_i(\lambda)},
\end{align}
as formalized by the theorem below. 
We focus on the underparameterized regime because only in that regime a linear estimator can have small risk for data generated from a linear model (with non-vanishing features). 
We consider the overparameterized regime in a more general setting empirically in the next section.

\begin{theorem}
    \label{thm:maindifference}
With probability at least
$1 - 2d^{-5} - 2de^{-n/8} - e^{-d} - 2e^{-32}$
over the random training set generated by a linear Gaussian model with parameters $\vtheta^\ast$ and  $\mSigma$,
the difference of$\,$ the $\ell_2$-regularized least squares risk and the risk expression in~\eqref{eq:defbarR} is at most 
\begin{align}
    \begin{split}
        & \left| \risk(\hat \vtheta_\lambda) - \bar R(\tilde \vtheta_\lambda) \right| \leq
        c
        \left[
            \frac{\max_i \sigma_i^8}{\min_i (\sigma_i^2 + \lambda)^4}
            \frac{d}{n}
            \quad \cdot
        \right.
        \\
        & \quad
        \left.
            \left(
                \left(\frac{\min_i \sigma_i^2 + \lambda}{\max_i \sigma_i^2} + 1\right) \norm[2]{\mSigma \vtheta^\ast} + \frac{d \log d}{\sqrt{n}} \sigma
            \right)^2
            +
            \frac{\sqrt{d}}{n} \sigma^2
        \right].
    \end{split}
\end{align}
Here, $c$ is a numerical constant.
\end{theorem}

Theorem~\ref{thm:maindifference} establishes that the risk $\risk(\hat \vtheta_\lambda)$ is well approximated by the expression $\bar R(\tilde \vtheta_\lambda)$, provided the model is sufficiently underparameterized (i.e., $d/n$ is small).

\begin{figure}

\begin{center}

\scalebox{0.95}{%
\begin{tikzpicture}[>=latex]

\draw[red,dashed,<-] (2.85cm,0cm) -- (2.85cm,3.4cm) node [above] {\small optimal regularization};

\draw[red,dashed,<-] (7.48cm,0cm) -- (7.48cm,3.4cm) node [above] {\small optimal regularization};

\draw[red,dotted] (2.85cm,0.96cm) -- (7.48cm,0.96cm);

\node at (2.85cm,1.45cm) [red,circle,fill,inner sep=1.5pt]{};
\node at (7.48cm,0.96cm) [red,circle,fill,inner sep=1.5pt]{};

\begin{groupplot}[
width=6.0cm,
height=4.8cm,
group style={
    group name=multi_dd,
    group size= 3 by 1,
    xlabels at=edge bottom,
    ylabels at=edge left,
    yticklabels at=edge left,
    xticklabels at=edge bottom,
    horizontal sep=0.8cm, 
    vertical sep=0.8cm,
    },
legend cell align={left},
legend columns=1,
legend style={
            at={(1,1)}, 
            anchor=north east,
            draw=black!30,
            fill=none,
            font=\scriptsize
        },
colormap name=viridis,
cycle list={[colors of colormap={50,350,750}]},
tick align=outside,
tick pos=left,
x grid style={white!69.01960784313725!black},
xtick style={color=black},
y grid style={white!69.01960784313725!black},
ylabel style={at={(-0.20,0.5)}},
ytick style={color=black},
xlabel={$1/\lambda$}, 
xmode=log,
title style={at={(0.5,-0.65)}},
]


\nextgroupplot[ylabel = {risk},title={\small a) constant $\lambda$},
            legend style={nodes={scale=0.72}}
]
      \addplot + [thick,mark=none,dash pattern=on 1pt off 2pt on 3pt off 2pt] table[x expr=1/\thisrow{lambda},y=risk1]{./fig/data/linear_risk_reconciled.dat};   
      \addlegendentry{bias-variance 1};
      \addplot + [thick,mark=none,dash pattern=on 1pt off 1pt on 1pt off 1pt] table[x expr=1/\thisrow{lambda},y=risk2]{./fig/data/linear_risk_reconciled.dat};
      \addlegendentry{bias-variance 2};
      \addplot + [thick,mark=none] table[x expr=1/\thisrow{lambda},y=risk_total]{./fig/data/linear_risk_reconciled.dat};
      \addlegendentry{1+2};


\nextgroupplot[title={\small b) elimination with diff. $\lambda$},
            legend style={nodes={scale=0.72}}
]
      \addplot + [thick,mark=none,dash pattern=on 1pt off 2pt on 3pt off 2pt] table[x expr=1/\thisrow{lambda},y=risk1]{./fig/data/linear_risk_reconciled.dat};   
      \addlegendentry{bias-variance 1};
      \addplot + [thick,mark=none,dash pattern=on 2pt off 3pt on 2pt off 1pt] table[x expr=1/\thisrow{lambda},y=new_risk2]{./fig/data/linear_risk_reconciled.dat};
      \addlegendentry{bias-variance 3};
      \addplot + [thick,mark=none] table[x expr=1/\thisrow{lambda},y=risk_total2]{./fig/data/linear_risk_reconciled.dat};
      \addlegendentry{1+3};

\nextgroupplot[title={\small c) before \& after elimination},
            legend style={nodes={scale=1.0}}
]

      \addplot + [thick,mark=none,green!70!black,dashed] table[x expr=1/\thisrow{lambda},y=risk_total]{./fig/data/linear_risk_reconciled.dat};
      \addlegendentry{1+2};
      
      \addplot + [thick,mark=none,green!70!black] table[x expr=1/\thisrow{lambda},y=risk_total2]{./fig/data/linear_risk_reconciled.dat};
      \addlegendentry{1+3};

\end{groupplot}

\end{tikzpicture}
}

\end{center}

\vspace{-0.2cm}

\caption{
\label{fig:mitigatingdouble}
Ridge regression risk for a two-feature Gaussian linear model as a function of the inverse regularization strength parameter $\lambda$.
{\bf a:}
Two U-shaped bias-variance tradeoffs $V_{i}(\lambda)$ 
for the parameters 
$\theta_1^\ast = 1.5, \sigma_1 = 1$ (bias-variance 1) and 
$\theta_2^\ast = 10, \sigma_2 = 0.15$
(bias-variance 2), along with their sum (1+2) which determines the risk. 
{\bf b:}
Same plot, but this time the bias-variance tradeoff $V_{2}(\lambda)$ is shifted to the left by increasing the inverse regularization strength $1/\lambda_2$ according to Proposition~\ref{prop:mincond} (yielding bias-variance tradeoff 3), so that its minimum overlaps with that of bias-variance tradeoff 1.
This eliminates double descent and gives better performance. 
{\bf c:} The resulting risk curves before and after elimination, demonstrating that the minimum of the risk after double descent elimination is smaller than before elimination.
}
\vspace{-0.25cm}
\end{figure}
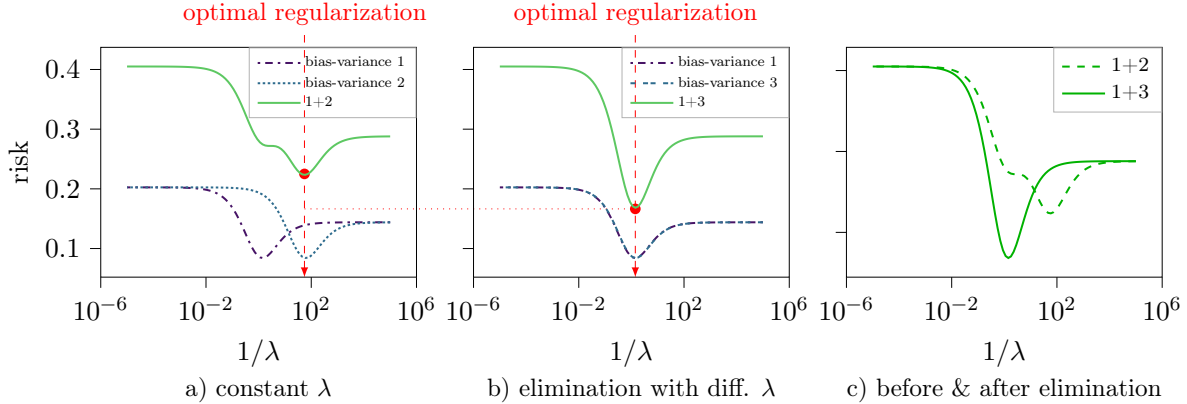

As a consequence, the risk of the ridge regression solution, as a function of $1/\lambda$, is a superposition of U-shaped bias variance tradeoffs. This yields double descent whenever the features of the underlying data have different scales. This follows from noting that the term $\sigma_i^2 (\theta_i^\ast)^2 \left(\frac{\lambda}{\sigma_i^2 + \lambda}\right)^2$ in the RHS of~\eqref{eq:defbarR} increases in $\lambda$, whereas the other term $\frac{\sigma^2}{n} \sigma_i^2 \left( \frac{\sigma_i}{\sigma_i^2 + \lambda} \right)^2$ decreases in $\lambda$. 
See Figure~\ref{fig:mitigatingdouble}(a) as an example.

\subsection{Eliminating double descent with scaled regularization}
\label{sec:linear_scaling_lambdas}

We next show that double descent can be eliminated by utilizing differently scaled $\lambda$ for different parts (parameters) of the model.
For this, we consider a generalized ridge regression problem where we allow different regularization strength to be used for each parameter (sometimes called Tikhonov regularization). Specifically, we let 
\begin{align}
\label{eq:thest}
\hat \vtheta_{\mLambda}
=
\arg \min_{\vtheta}
\frac{1}{2} 
\sum_{i=1}^n (\innerprod{\vx_i}{\vtheta} - y_i)^2
+ \norm[2]{\mLambda \vtheta}^2,
\end{align}
where $\mLambda$ is a $\reals^{d \times d}$ diagonal matrix containing regularization parameters $\sqrt{\lambda_i}$ along its diagonal.

\begin{proposition}
\label{prop:mincond}
For the generalized ridge regression problem described above, the minimum of the risk expression $\min_{\lambda_1, \ldots, \lambda_d} \bar \risk (\tilde \theta_{\mLambda})$ is achieved by choosing the regularization strengths associated with different features as $\lambda_i = \frac{\sigma^2}{n} (\theta_i^\ast)^{-2}$.
\end{proposition}

In Figure~\ref{fig:mitigatingdouble}b, we show that double descent can be eliminated, and that this improves the optimal risk, by utilizing the regularization parameters in Proposition~\ref{prop:mincond}. Note that double descent is eliminated by picking the optimal regularization strength associated with feature $j$ as $\bar \lambda_{opt} = \lambda_j = \frac{\sigma^2}{n} (\theta_j^\ast)^{-2}$ and scaling the regularization strengths of the rest of the features proportionally with $(\theta_j^\ast / \theta_i^\ast)^2$ to align the minima of the U-shaped bias-variance tradeoff curve $V_i(\lambda_i)$ with the minima of the bias-variance tradeoff curve of the $j^{th}$ feature $V(\bar \lambda_{opt})$.

Note that the optimal regularization strength does not depend on the variances of the features, i.e., at the optimal regularization point, the effect of the feature variances on the bias and variance components of the risk is equal in magnitude, i.e., the tradeoff does not depend on the feature variances other than a constant scaling factor for the both bias and variance terms (see SM~\ref{app:proof_prop1}).

\subsection{Relation to early stopping}
\label{sec:relation2early_stopping}

As already illustrated in Figure~\ref{fig:mcnn_lambda}, in general, $\ell_2$ regularization and early stopping have a different effect. 
However, they
\emph{can} have a similar~\citep{ali_ContinuousTimeViewEarly_2019}, and  even equivalent effect in very particular setups. 
For example, for the linear model studied so far, Tikhonov regularization and early stopping has the same effect if we  adjust the regularization strength parameters associated with individual parameters.

Consider the Tikhonov estimator defined in~\eqref{eq:thest}. 
Also consider the estimator which applies $t$ steps of gradient descent to the non-regularized loss $\sum_{i=1}^n (\innerprod{\vx_i}{\vtheta} - y_i)^2$, and suppose that each parameter $\theta_i$ is updated with an associated stepsize of $\eta_i$. This estimator, denoted by $\hat \vtheta^t$ corresponds to early-stopping least-squares. 
This estimator was studied  by~\citet{heckel_EarlyStoppingDeep_2020} and shown to have risk 
\begin{align}
    \label{eq:risk_early_stopping}
    R(\hat \vtheta^t)
    \approx
    \sigma^2 +
    \sum_{i=1}^d 
    \underbrace{
    \sigma_i^2
    (\theta^\ast_i)^2
    (1 - \eta_i \sigma_i^2)^{2\iter}  
    +  
    \frac{\sigma^2}{n} (1 - (1-\eta_i \sigma_i^2)^\iter)^2
    }_{U_i(\iter)}.
\end{align}
As formalized by the following proposition, if $\lambda_i$ are chosen based on the feature variance $\sigma_i$ and the corresponding stepsize $\eta_i$, then the risk expressions for the corresponding Tikhonov estimator and early-stopped least squares are equivalent:

\begin{proposition}
\label{lem:lambda2earlystopping}
    Let $\mLambda = \diag(\sqrt{\lambda_1},\ldots,\sqrt{\lambda_d})$ 
    with $\lambda_i = \frac{\sigma_i^2}{1 - (1 - \eta_i \sigma_i^2)^{\iter}} -\sigma_i^2$. Then the risk of Tikhonov regularized least-squares is equal to the risk of early-stopping the gradient descent iterations applied to the non-regularized loss at time $t$ as given in equation~\eqref{eq:risk_early_stopping}. 
\end{proposition}

The above proposition characterizes the requirement such that the bias variance tradeoff curves induced by regularized-least squares are equivalent when using $\ell_2$ regularization or regularization by early stopping.
However, note that this requires the regularization parameters $\lambda_i$ to be dependent on the feature variances $\sigma_i^2$.  
In general, where the regularization parameters and stepsizes are the same for each parameter, the risk corresponding to regularization by early stopping and $\ell_2$ regularization is different.

\section{Double descent in $\ell_2$-regularized two-layer neural networks}
\label{sec:two_layer_ridge}

In this section, we study the risk of a two layer network with weight decay (i.e., $\ell_2$-penalty), on data drawn from a Gaussian linear model with a diagonal covariance matrix.
We first show empirically that the risk as a function of the regularization parameter has a double descent curve if the variances of the Gaussian model's features decay at a geometric rate, and that the double descent can be eliminated by penalizing the weights in the first and second layers differently.

While it would be nice to explain this theoretically, this is not possible with current linearization techniques: we show that regularization-wise double descent in neural networks occurs outside of the regime where the network dynamics can be characterized by an associated linear model (often called the neural-tangent-kernel (NTK) regime.

\subsection{Risk of an overparameterized two-layer network exhibits double descent}

We consider a two-layer neural network with relu-nonlinearities, $f(\vx) = \relu(\mW_1 \vx) \vw_2$, where $\mW_1 \in \reals^{k\times d}$ and $\vw_2\in \reals^k$ are the weights in the first and second layer. 
The network is trained with gradient descent on the mean-squared error loss with $\ell_2$-penalty on data drawn from the linear model introduced in Section~\ref{sec:datamodel} with a diagonal covariance matrix with geometrically decaying covariances and Gaussian zero-mean additive noise.
For each value of the regularization parameter $\lambda$, we initialize the network with standard Kaiming initialization and train until convergence with stepsize $\eta = 5e-3$.

\begin{figure}[t]

\begin{center}

\scalebox{0.95}{%
\begin{tikzpicture}[>=latex]

\begin{groupplot}[
width=7cm,
height=5cm,
group style={
    group name=two_layer_regularization,
    group size= 2 by 1,
    xlabels at=edge bottom,
    ylabels at=edge left,
    xticklabels at=edge bottom,
    horizontal sep=2cm, 
    vertical sep=1.8cm,
    },
legend cell align={left},
legend columns=1,
legend style={
            at={(1,1)}, 
            anchor=north east,
            draw=black!50,
            fill=none,
            font=\small
        },
colormap name=viridis,
cycle list={[colors of colormap={50,250,450,700}]},
tick align=outside,
tick pos=left,
x grid style={white!69.01960784313725!black},
xtick style={color=black},
y grid style={white!69.01960784313725!black},
ytick style={color=black},
xlabel={$\iter$ iterations}, 
ylabel={risk},
xmode=log,
ymin=9.5, ymax=12,
every axis plot/.append style={thick}
]


\nextgroupplot[xlabel={$1/\lambda$},
            legend style={
                nodes={scale=1.0, transform shape},
                at={(1,1)}, 
                anchor=north east,
                draw=black!50,
                fill=none,
                font=\small
            },
        ]
    
    \addplot +[mark=none] table[x expr=1/\thisrow{lambda},y=risk,select coords between index={0}{45}]{./fig/data/two_layer_nn_lambda_risk_scaled.txt};
    \addlegendentry{same $\lambda$}

    \addplot +[mark=none,dash pattern=on 2pt off 1pt] table[x expr=1/\thisrow{lambda},y=scaled,select coords between index={0}{42}]{./fig/data/two_layer_nn_lambda_risk_scaled.txt};
    \addlegendentry{scaled $\lambda$}

\nextgroupplot[]

    \addplot +[mark=none] table[x=t,y=risk,select coords between index={5}{1000}]{./fig/data/two_layer_nn_t_risk.txt};

\end{groupplot}

\end{tikzpicture}
}

\end{center}

\vspace{-0.2cm}

\caption{
\label{fig:twolayerregularization}
{\bf Left:} Risk of the two-layer neural network trained on the linear data with a diagonal covariance matrix with geometrically decaying singular values and added noise with $\ell_2$ regularization as a function of the inverse regularization parameter $1/\lambda$. The risk exhibits the double descent behavior.
{\bf Right:}
Same risk as a function of the training iterations $\iter$ for $\lambda = 0$. The early-stopped risk does not yield a double descent behavior.
{\bf Both:} Training dynamics of regularization by early stopping cannot be approximated by the solutions of the corresponding $\ell_2$ regularization problem.
}
\vspace{-0.4cm}
\end{figure}
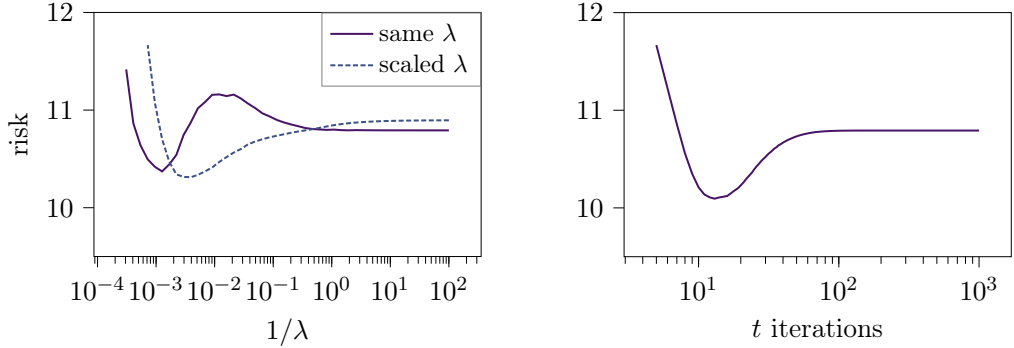

Figure~\ref{fig:twolayerregularization} shows that the resulting risk 
follows a double descent curve as a function of $1/\lambda$. 
Figure~\ref{fig:twolayerregularization} also shows that the risk of early-stopped gradient descent, while 
operating in the same range of values,
does not exhibit double descent. 
This again illustrates that $\ell_2$ regularization and regularization by early stopping in general result in different risk curves, as formalized in the previous section for linear models.

Recall that double descent for linear models occurs because different features are scaled differently, and can be mitigated by 
scaling $\lambda_i$ appropriately,
as formalized in Proposition~\ref{prop:mincond} and demonstrated in Figure~\ref{fig:mitigatingdouble}. 
Motivated by this result, we hypothesize that the first and second layers of the two-layer neural network overfit the noise at different scales. 
Thus, utilizing properly scaled $\lambda_1$ and $\lambda_2$ for the parameters in the first and second layers should mitigate double descent and potentially improve performance. 

In Figure~\ref{fig:twolayerregularization}, we show that 
double descent is indeed eliminated by using a larger $\lambda$ for the second layer and that the best performance (i.e., the performance achieved at the optimal regularization point) is improved relative to the best performance for the risk curve where double descent is not eliminated. 
Note that in the linear case studied in Section~\ref{sec:linear_scaling_lambdas}, when the parameters of the underlying data model are known or can be estimated, the optimal $\lambda_i$, i.e. per feature regularization strength, can be found analytically. In contrast, for neural networks, this requires treating per-layer regularization strengths as hyperparameters and tuning them accordingly.


\subsection{Double descent occurs outside the linear regime in neural networks}
\label{sec:noNTK}

Given our theoretical results for the linear model, and the similar empirical behavior of linear models and neural networks, it is tempting to think that the behavior of the two-layer network from the previous section (and potentially deeper networks) can be described theoretically by linearizing the network around the initialization, and studying the linearized model as a proxy for the actual non-linear network. 
This regime is known as the NTK regime~\citep{jacot_NeuralTangentKernel_2018} 
because the model behaves like a kernel method with a kernel associated with the neural network called neural tangent kernel. 

Unfortunately, double descent as a function of $\lambda$ occurs outside of the regime where a linear approximation is accurate, as we discuss here. 

Consider a neural network with parameter vector $\vtheta$ and input $\vx$, denoted by $f_\vtheta(\vx)$. Suppose we train the network on a dataset $\{(\vx_1,y_1),\ldots,(\vx_n,y_n) \}$ by applying gradient descent to the $\ell_2$-regularized least-squares loss 
\begin{align*}
\mc L_\lambda(\vtheta)
=
\frac{1}{2} \sum_{i=1}^n (f_{\vtheta}(\vx_i) - y_i )^2 + \frac{\lambda}{2} \norm[2]{\vtheta}^2
\end{align*}
until convergence. 
The predictions of the network in a small radius around the initialization $\vtheta_0$ are well described by the linear approximation $ \vf_{\vtheta} \approx
        \mJ
        (\vtheta  - \vtheta_0) + \vf_{\vtheta_0}$,
where 
\begin{align}
        \vf_\vtheta
        =
        \begin{bmatrix}
        f_{\vtheta}(\vx_1) \\
        \hdots
        \\
        f_{\vtheta}(\vx_n) 
        \end{bmatrix}%
        \;
        \text{and}
        \;
        \mJ
        =
        \begin{bmatrix}
        \nabla_\vtheta f_{\vtheta}(\vx_1) \\
        \hdots
        \\
        \nabla_\vtheta f_{\vtheta}(\vx_n) 
        \end{bmatrix}
\end{align}
are the prediction of the network and the Jacobian of the network at initialization, respectively. 
The linear approximation is only accurate in a radius around the initialization, in which each individual parameter changes very little. 
However, as we argue in more detail in the supplement, the individual parameters change too much for this approximation to be accurate (see Figure~\ref{fig:twolayergradients}, left), unless the singular values of the Jacobian are large relative to $\lambda$.
However, we note that the individual parameters change too much for this approximation to be accurate, unless the singular values of the Jacobian are large relative to $\lambda$.
If the singular values are sufficiently large for the NTK approximation to be accurate, however, the regularization has a vanishing effect, and in the regime where the regularization has a vanishing effect, no double descent occurs. We refer to SM~\ref{app:noNTK} for a more detailed analytical discussion.

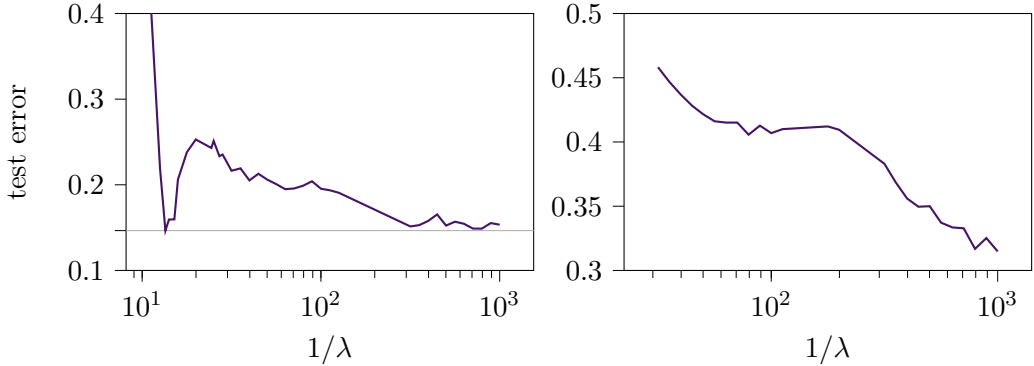
\begin{figure}[t]
    \begin{center}
    \begin{tikzpicture}
    
    \begin{groupplot}[
    width=7cm,
    height=5cm,
    group style={
        group name=mcnn_regularization,
        group size=3 by 1,
        xlabels at=edge bottom,
        ylabels at=edge left,
        xticklabels at=edge bottom,
        horizontal sep=1.2cm, 
        vertical sep=1.8cm,
        },
    legend cell align={left},
    legend columns=1,
    legend style={
                at={(1,1)}, 
                anchor=south east,
                draw=black!50,
                fill=none,
                font=\small
            },
    /tikz/every even column/.append style={column sep=0.1cm},
    /tikz/every row/.append style={row sep=0.3cm},
    colormap name=viridis,
    cycle list={[colors of colormap={50,350,750}]},
    tick align=outside,
    tick pos=left,
    x grid style={white!69.01960784313725!black},
    xtick style={color=black},
    y grid style={white!69.01960784313725!black},
    ytick style={color=black}, 
    extra y ticks={0.1465}, 
    extra y tick style={%
        grid=major,
    },
    extra y tick labels=,
    xlabel={$1/\lambda$}, 
    ylabel={test error},
    ymin=0.1, ymax=0.4,
    xmode=log,
    every axis plot/.append style={thick},
    ]

    
    \nextgroupplot[]
    
    \addplot +[mark=none] table[x expr=1/\thisrow{lambda},y expr=1 - \thisrow{test}/100]{./fig/data/resnet_lambda_error_smooth.txt};

    
    \nextgroupplot[ymin=0.3, ymax=0.5] 
    
    \addplot +[mark=none] table[x expr=1/\thisrow{lambda},y expr=1 - \thisrow{test}/100]{./fig/data/mcnn_cifar100_lambda_approximate_dd.txt};
    
    \end{groupplot}
    
    \end{tikzpicture}
    \end{center}
    \vspace{-0.2cm}
    \caption{\label{fig:resnet_cifar10_mcnn_cifar100}
    Regularization-wise double descent for models and datasets of more practical interest: 
    {\bf (Left)} Test performance of ResNet-18 as a function of the inverse regularization parameter ($1/\lambda$) when trained on the CIFAR-10 dataset with 20\% label noise exhibits double descent;
    {\bf (Right)} Test performance of the 5-layer CNN as a function of the inverse regularization parameter ($1/\lambda$) when trained on the CIFAR-100 dataset with \emph{no label noise} also exhibits (subtler) double descent
   }
   \vspace{-0.3cm}
\end{figure}

\section{Double descent in deep networks}
\label{sec:deep_dd}

We next study a 5-layer CNN and ResNet-18 to demonstrate that regularization-wise double descent occurs in standard deep learning settings.
We first look at the test error of a 5-layer CNN trained on the CIFAR-10 dataset with 20\% label noise as a function of the regularization strength, or weight decay.
We also compare this curve to the unregularized training curve as a function of training epochs, which also exhibits double descent, to demonstrate that the two regularizations function distinctively differently. 

Our results in Figure~\ref{fig:mcnn_lambda} show that the test error as a function of regularization strength follows a double descent curve.
Moreover, 
while there is a clear optimal $\lambda$ value where the minimum test error is achieved in the small $\lambda$ regime, which coincides with the typical values of weight decay used in practice, a similar performance can be achieved in the much larger $\lambda$ regime. 

Note that double descent can be potentially eliminated with \emph{more} regularization.~\citet{nakkiran_OptimalRegularizationCan_2020} showed that sample-wise double descent can be eliminated by employing optimal $\ell_2$-regularization.
We report that regularization-wise double descent can also be eliminated by employing early-stopping in conjunction with weight decay and epoch-wise double descent by employing optimally-tuned weight decay (see SM~\ref{app:deep_wd_es}, Figure~\ref{fig:mcnn_wd_es}).

Moreover, in both cases, eliminating the double descent improves the performance compared to the case where $\ell_2$ regularization or early stopping is individually applied.



While CNNs are commonly used for vision applications, standard architectures feature more complex mechanisms, such as residual links, and hence the training dynamics of such models can significantly vary from that of the simple 5-layer CNN.
We therefore also study the test error of the ResNet-18 model trained on the CIFAR-10 dataset with 20\% label noise.
We show that, in Figure~\ref{fig:resnet_cifar10_mcnn_cifar100} (left), the test error for ResNet-18 also exhibits double descent even though the achieved performance across all $\lambda$ values is better for ResNet-18 than the 5-layer CNN as can be expected.
Moreover, similarly to the case of the 5-layer CNN, a similar test error can be achieved at both small and large $\lambda$ regimes.

For deep learning models trained on image classification datasets, the double descent phenomenon is primarily observed when the model is trained on noisy data. For example, epoch-wise double descent~\citep{nakkiran_DeepDoubleDescent_2020,heckel_EarlyStoppingDeep_2020} has only been observed in practical setups when training on noisy data (i.e., data with label noise). 

We next show that regularization-wise double descent can also occur in more practical settings, i.e. when there is no label noise, which is the most common situation in practice. 
Our results in Figure~\ref{fig:resnet_cifar10_mcnn_cifar100} (right) show that the test error of the 5-layer CNN trained on the CIFAR-100 dataset with no label noise also exhibits double descent, albeit in a less pronounced manner. This is expected, since higher levels of noise in general lead to a more pronounced double descent curve. 

\section{Conclusion}
In this work, we studied regularization-wise double descent in an effort to bring its understanding to the same level as the previously well-studied model-wise, epoch-wise and sample-wise double descents. 
We demonstrated that the test error of standard deep networks trained on standard image classification datasets can follow a double descent curve as a function of $\ell_2$ regularization strength (weight decay) both when there is label noise (CIFAR-10) and without any label noise (CIFAR-100). 

We show that regularization-wise double descent can be explained as a superposition of bias-variance tradeoffs pertaining to different features of the data (for a linear model) or parts of the neural network, and that double descent can be eliminated by scaling the regularization strengths accordingly. 



\section*{Code}
Code to reproduce the experiments is available at 
\url{https://github.com/MLI-lab/regularization-wise_double_descent}.

\section*{Acknowledgements}
F. F. Yilmaz and R. Heckel are (partially) supported by NSF under award IIS-1816986. R. Heckel is also supported by the Institute of Advanced Studies at the Technical University of Munich, and the Deutsche Forschungsgemeinschaft (DFG, German Research Foundation) - 456465471, 464123524.

\clearpage
\printbibliography

\clearpage
\appendix

\section{Double descent behavior of deep networks in the presence of both weight decay and early stopping}
\label{app:deep_wd_es}
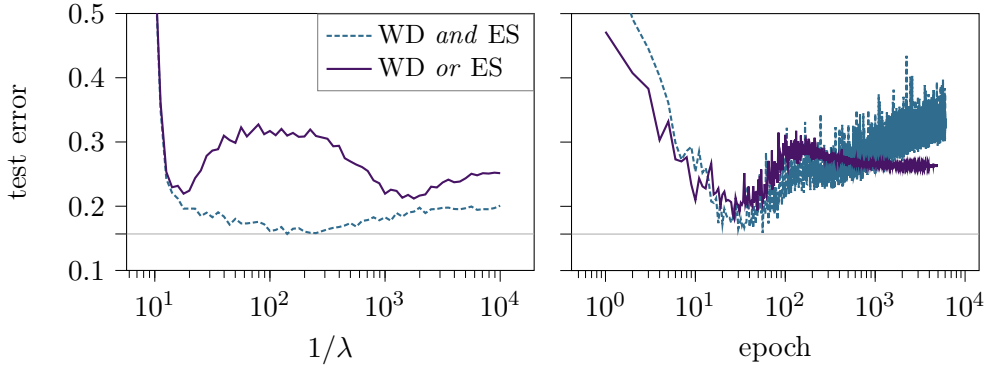
\begin{figure}[t]
    \begin{center}
    \begin{tikzpicture}
    
    \begin{groupplot}[
    width=7cm,
    height=5cm,
    group style={
        group name=mcnn_regularization,
        group size=3 by 1,
        xlabels at=edge bottom,
        ylabels at=edge left,
        yticklabels at=edge left,
        xticklabels at=edge bottom,
        horizontal sep=0.5cm, 
        vertical sep=1.8cm,
        },
    legend cell align={left},
    legend columns=1,
    legend style={
                at={(1,1)}, 
                anchor=north east,
                draw=black!50,
                fill=none,
                font=\small
            },
    /tikz/every even column/.append style={column sep=0.1cm},
    /tikz/every row/.append style={row sep=0.3cm},
    colormap name=viridis,
    cycle list={[colors of colormap={350,50,750}]},
    tick align=outside,
    tick pos=left,
    x grid style={white!69.01960784313725!black},
    xtick style={color=black},
    y grid style={white!69.01960784313725!black},
    ytick style={color=black},
    extra y ticks={0.1568}, 
    extra y tick style={%
        grid=major,
    },
    extra y tick labels=,
    xlabel={$1/\lambda$}, 
    ylabel={test error},
    ymin=0.1, ymax=0.5,
    xmode=log,
    every axis plot/.append style={thick},
    ]
    
    
    \nextgroupplot[]

    \addplot +[mark=none,dash pattern=on 2pt off 1pt] table[x expr=1/\thisrow{lambda},y expr=1 - \thisrow{test}/100]{./fig/data/mcnn_WD_and_ES.txt};
    \addlegendentry{WD \emph{and} ES}
    
    \addplot +[mark=none] table[x expr=1/\thisrow{lambda},y expr=1 - \thisrow{test}/100]{./fig/data/mcnn_lambda_error_extended.txt};
    \addlegendentry{WD \emph{or} ES}

    \nextgroupplot[xlabel=epoch]

    \addplot +[mark=none,dash pattern=on 2pt off 1pt] table[x=epoch,y expr=1 - \thisrow{test}/100]{./fig/data/mcnn_wd_es_no_epoch_dd.txt};
    
    \addplot +[mark=none] table[x=epoch,y expr=1 - \thisrow{test}/100]{./fig/data/mcnn_double_descent.txt};

    \end{groupplot}
    
    \end{tikzpicture}
    \end{center}
    \vspace{-0.2cm}
    \caption{\label{fig:mcnn_wd_es}
    Comparison of individually or jointly applied regularization by early stopping and weight decay: 
    Test performance of the 5-layer convolution network when trained on the CIFAR-10 dataset with 20\% label noise.
    {\bf Left:}
    Performance as a function of the regularization strength for training with (\emph{solid}) weight decay only---WD---and (\emph{dashed}) weight decay together with early stopping---WD and ES.
    {\bf Left:}
    Performance as a function of the training epochs for (\emph{solid})  standard training and (\emph{dashed}) training with weight decay.
    {\bf Both:} 
    Better performance is achieved by jointly utilizing weight decay and early stopping---WD and ES.
    }
\end{figure}

Here, we expand on the results provided in Figure~\ref{fig:mcnn_lambda} and show that both regularization-wise and epoch-wise double descent can be eliminated by employing additional forms of regularization.
Specifically, in Figure~\ref{fig:mcnn_wd_es}, our results show that utilizing early stopping eliminates regularization-wise double descent, whereas utilizing (tuned) weight decay eliminates the corresponding epoch-wise double descent.
Note that performance achieved in the case where early stopping and weight decay are used together is much better than that obtained by using either weight decay or early stopping alone.

\section{Double descent as a function of dropout regularization}
\label{app:dropout}

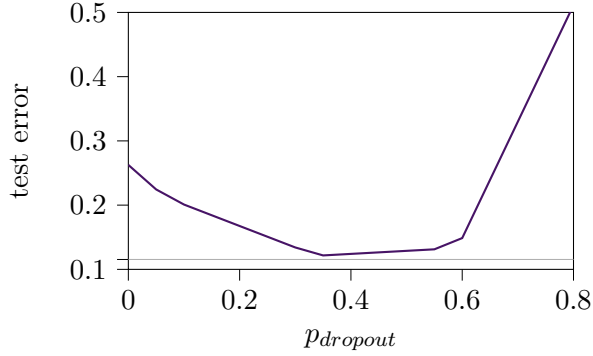
\begin{figure}[t]
    \begin{center}
    \begin{tikzpicture}
    
    \begin{groupplot}[
    width=7.5cm,
    height=5cm,
    group style={
        group name=mcnn_regularization,
        group size=3 by 1,
        xlabels at=edge bottom,
        ylabels at=edge left,
        yticklabels at=edge left,
        xticklabels at=edge bottom,
        horizontal sep=0.5cm, 
        vertical sep=1.8cm,
        },
    legend cell align={left},
    legend columns=1,
    legend style={
                at={(1,1)}, 
                anchor=south east,
                draw=black!50,
                fill=none,
                font=\small
            },
    /tikz/every even column/.append style={column sep=0.1cm},
    /tikz/every row/.append style={row sep=0.3cm},
    colormap name=viridis,
    cycle list={[colors of colormap={50,350,750}]},
    tick align=outside,
    tick pos=left,
    x grid style={white!69.01960784313725!black},
    xtick style={color=black},
    y grid style={white!69.01960784313725!black},
    ytick style={color=black},
    extra y ticks={0.1154}, 
    extra y tick style={%
        grid=major,
    },
    extra y tick labels=,
    xlabel={$p_{dropout}$}, 
    ylabel={test error},
    ymin=0.1, ymax=0.5,
    xmin=0, xmax=0.8,
    every axis plot/.append style={thick},
    ]
    
    
    \nextgroupplot[]
    
    \addplot +[mark=none] table[x expr=\thisrow{dropout},y expr=1 - \thisrow{test}/100]{./fig/data/mcnn_dropout_error.txt};

    \end{groupplot}
    
    \end{tikzpicture}
    \end{center}
    \vspace{-0.2cm}
    \caption{\label{fig:mcnn_dropout}
    Test performance of the 5-layer convolution network as a function of the dropout probability when trained on the CIFAR-10 dataset with 20\% label noise.
    }
\end{figure}

Our results showcasing the double descent behavior as a function of the $\ell_2$ regularization strength motivates the investigation of other types of regularization and whether double descent also occurs for other explicit regularization methods. In Figure~\ref{fig:mcnn_dropout}, we show the test error of the 5-layer CNN with dropout added after the activations of each layer trained on the noisy CIFAR-10. The test error exhibits a U-shaped curve  as a function of the dropout probability with optimal dropout probability $p_{dropout} = 0.4$.

\section{Discussion and proof statements for linear ridge regression}
\label{app:linear_ridge_regression}
In this section, we provide detailed analysis and proofs for the theoretical statements on the linear ridge regression risk studied in Section~\ref{sec:ridgeregression}.

\subsection{Intuition for the risk expression~\eqref{eq:defbarR}}
We first provide intuition on why the risk is governed by the risk expression given in~\eqref{eq:defbarR}. 

First, note that the risk of the resulting estimator can be written as a function of the variances of the features, $\sigma_i^2$, and of the coefficients of the underlying true linear model, $\vtheta^\ast = [\theta_1^\ast,\ldots, \theta_d^\ast]$, as
\begin{align}
\risk(\hat \vtheta_\lambda)
&=
\sigma^2
+
\sum_{i=1}^d \sigma_i^2 (\theta^\ast_i - \hat \theta_{\lambda, i})^2.
\label{eq:expressionlambda}
\end{align}
which follows from noting that $z$ and $\vx$ are independently drawn.

Next, note that we aim to find the estimator which minimizes the $ell_2$-regularized MSE loss
\begin{align*}
\mc L_\lambda(\vtheta) = \frac{1}{2} \norm[2]{ \mX \vtheta - \vy }^2 + \frac{\lambda}{2} \norm[2]{\vtheta}^2.
\end{align*}

Recall that, as introduced in Section~\ref{sec:datamodel} , the matrix $\mX \in \reals^{n\times d}$ contains the scaled training feature vectors $\frac{1}{\sqrt{n}}\vx_1,\ldots, \frac{1}{\sqrt{n}} \vx_n$ as rows, and $\vy = \frac{1}{\sqrt{n}}[y_1,\ldots,y_n]$ are the corresponding scaled responses. 
Then, the solution of the $\ell_2$ regularized problem can be found by simply setting the gradient of the loss function to zero and solving for $\vtheta$, which yields
\begin{align*}
    \vtheta_\lambda - \vtheta^\ast 
    &=
    ((\transp{\mX} \mX + \lambda \mI)^{-1} \transp{\mX} \mX - \mI) \vtheta^\ast 
    +
    (\transp{\mX} \mX + \lambda \mI)^{-1} \transp{\mX}\vz,
\end{align*}
where $\vz = [z_1,\ldots, z_n]$ is the noise. 
As we formalize below, in the under-parameterized regime where $n \gg d$, we have that $\transp{\mX} \mX \approx \mSigma^2$. Therefore the original solution is close to the proximal solution $\tilde \vtheta_\lambda$ defined by
\begin{align}
    \label{eq:closesolutions}
    \tilde \vtheta_\lambda - \vtheta^\ast 
    &=
    ((\transp{\mSigma} \mSigma + \lambda \mI)^{-1} \transp{\mSigma} \mSigma - \mI) \vtheta^\ast 
    +
    (\transp{\mSigma} \mSigma + \lambda \mI)^{-1} \transp{\mX}\vz,
\end{align}
The proximal solution is close to the original solution obtained by solving for the minimizer of the $\ell_2$-regularized loss function.
Note that, from~\eqref{eq:closesolutions}, we get, for the i-th entry of $\tilde \vtheta_\lambda$
\begin{align*}
    \tilde \vtheta_{\lambda, i} - \vtheta^\ast_i
    &=
    \transp{\tilde \vx_i} \vz \frac{1}{\sigma_i^2 + \lambda}
    -
    \frac{\lambda}{\sigma_i^2 + \lambda} \vtheta^\ast_i,
\end{align*}
where $\tilde \vx_i$ is the $i$-th \emph{column} of $\mX$ (not the $i$-th example/feature vector!). 
Next note that, $\EX{ (\transp{\tilde \vx}_i\vz)^2 } \approx \sigma^2 \sigma_i^2$ 
because the entries of $\vz$ are $\mc N(0,\sigma^2)$ distributed, and the entries of $\tilde \vx_i$ are $1/\sqrt{n} \mc N(0,\sigma_i^2)$ distributed. 
Using this expectation in the solution $\tilde \vtheta_\lambda$, and evaluating the resulting risk of those iterates via the formula for the risk given by~\eqref{eq:expressionlambda} yields the risk expression~\eqref{eq:defbarR}. 
The proof of Theorem~\ref{thm:maindifference} in this appendix makes this intuition precise by formally bounding the difference of the proximal solution $\tilde \vtheta_\lambda$ to the original solution $\vtheta_\lambda$. 


\subsection{Proof of Theorem~\ref{thm:maindifference}}
\label{app:proof_theorem_1}
In this section, we provide the formal proof for Theorem~\ref{thm:maindifference}.

The difference between the two risk terms can be further dissected into two separate terms:

\begin{align}
\label{eq:stdbound}
\left| \risk(\vtheta_\lambda) - \bar R(\tilde \vtheta_\lambda) \right|
&\leq
\left| \risk(\vtheta_\lambda) - \risk(\tilde \vtheta_\lambda) \right|
+
\left| 
\risk(\tilde \vtheta_\lambda) - \bar R(\tilde \vtheta_\lambda)
\right|.
\end{align}

We bound the two terms on the RHS of~\eqref{eq:stdbound} separately. We first provide a bound for the first term with the lemma below.

\begin{lemma}
    \label{lem:boundlambdas}
Define $\tilde \mX$ so that $\mX = \tilde \mX \mSigma$.  
Suppose that $\norm{ \mI - \transp{\tilde \mX} \tilde \mX} \leq \epsilon$, with 
$\epsilon \leq (\min_i \sigma_i^2 + \lambda)/2$
Then

\begin{align}
\label{eq:boundlambdas}
\left| \risk(\vtheta_\lambda) - \risk(\tilde \vtheta_\lambda) \right|
\leq
4 \epsilon^2
\left(
    \frac{\max_i \sigma_i^4}{\min_i (\sigma_i^2 + \lambda)^2}
\right)^2
\left(
    \left(\frac{\min_i \sigma_i^2 + \lambda}{\max_i \sigma_i^2} + 1\right) \norm[2]{\mSigma \vtheta^\ast} + \norm[2]{\transp{\tilde \mX} \vz}
\right)^2
%
%
\end{align}

\end{lemma}

We apply the lemma by first verifying its condition by referring to the derivations in~\cite[Lemma 1]{heckel_EarlyStoppingDeep_2020}. Note that the entries of the matrix $\tilde \mX$ are iid Gaussians drawn from $\mc N(0,1/n)$, and the same concentration inequality from~\cite[Chapter 9]{foucart_MathematicalIntroductionCompressive_2013} results in, for any $\beta \in (0,1)$,

\begin{align*}
\PR{ \norm{ \mI - \transp{\tilde \mX} \tilde \mX } \geq \beta }
\leq e^{- \frac{n\beta^2}{15} + 4d}.
\end{align*}
With $\beta  = \sqrt{\frac{75d}{n}}$ we obtain that, with probability at least $1-e^{-d}$, 
\begin{align*}
 \norm{ \mI - \transp{\tilde \mX} \tilde \mX } \leq \sqrt{75\frac{d}{n}}.
\end{align*}

We next bound $\norm[2]{\transp{\tilde \mX} \vz}$ with high probability:
\begin{lemma}
    \label{lem:normTildeXTz}
    With $\tilde \mX$ previously defined such that $\mX = \tilde \mX \mSigma$, with probability at least $1 - 2d(e^{-\beta^2/2} + e^{-n/8})$,
    \begin{align*}
        \norm[2]{\transp{\tilde \mX} \vz}
        \leq
        2 \frac{d}{\sqrt{n}} \sigma \beta
    \end{align*}
\end{lemma}

Applying the lemma with $\beta^2 = 10\log(d)$, we obtain that with probability at least
$1 - 2d^{-5} - 2de^{-n/8} - e^{-d}$ we have

\begin{align*}
    \left| \risk(\vtheta_\lambda) - \risk(\tilde \vtheta_\lambda) \right|
    \leq
    4 \frac{75d}{n}
    \left(
        \frac{\max_i \sigma_i^4}{\min_i (\sigma_i^2 + \lambda)^2}
    \right)^2
    \left(
        \left(\frac{\min_i \sigma_i^2 + \lambda}{\max_i \sigma_i^2} + 1\right) \norm[2]{\mSigma \vtheta^\ast} + 2 \frac{d}{\sqrt{n}} \sigma 10 \log d
    \right)^2
\end{align*}

We finally bound the second term in~\eqref{eq:stdbound}:

\begin{lemma}
    \label{lem:riskvsexp}
Provided that $d/n \leq \max_i ((\sigma_i + \lambda) / \sigma_i^2)^4$,
with probability at least $1 - 4e^{- \frac{\beta^2}{8}}$, we have that
\begin{align}
\label{eq:riskvsexp}
\left| 
\risk(\tilde \vtheta_\lambda) - \bar R(\tilde \vtheta_\lambda)
\right|
\leq
\frac{\sigma^2}{n} \beta
3 \sqrt{d},
\end{align}
with $\bar R(\tilde \vtheta_\lambda)$ as defined in~\eqref{eq:defbarR}. 
\end{lemma}
For the proof of Lemma~\ref{lem:riskvsexp} we refer the reader to the proof of~\cite[Lemma 2]{heckel_EarlyStoppingDeep_2020} and note that~\eqref{lem:riskvsexp} can be obtained by following the same steps with the additional assumption regarding the underparameterization as stated in Lemma~\ref{lem:riskvsexp}.

We note that the assumption of the lemma is generally satisfied as we operate in the underparameterized regime and poses no strict restriction on the setup.
Applying the two bounds~\eqref{eq:boundlambdas} and~\eqref{eq:riskvsexp} to the RHS of the bound~\eqref{eq:stdbound} concludes the proof. 
The remainder of the proof is devoted to proving Lemma~\ref{lem:boundlambdas}.


\subsection{Proof of Lemma~\ref{lem:boundlambdas}}

Recall that the solutions of the original and closely related problem are given by
\begin{align*}
\vtheta_\lambda - \vtheta^\ast 
&=
((\transp{\mX} \mX + \lambda \mI)^{-1} \transp{\mX} \mX - \mI) \vtheta^\ast 
+
(\transp{\mX} \mX + \lambda \mI)^{-1} \transp{\mX}\vz,
\\
\tilde \vtheta_\lambda - \vtheta^\ast 
&=
((\mSigma^2 + \lambda \mI)^{-1} \mSigma^2 - \mI) \vtheta^\ast 
+
(\mSigma^2 + \lambda \mI)^{-1} \transp{\mX}\vz.
\end{align*}

Note that $\mX = \tilde \mX \mSigma$, where we defined $\tilde \mX$ which has iid Gaussian entries $\mc N(0,1/n)$.
With this notation, and using that $\mSigma$ is diagonal and therefore commutes with symmetric matrices, we obtain the following expressions for the residuals of the two solutions:
\begin{align*}
\mSigma \vtheta_\lambda - \mSigma \vtheta^\ast 
&=
\mSigma ((\mSigma^2 \transp{\tilde \mX} \tilde \mX + \lambda \mI)^{-1} \mSigma^2 \transp{\tilde \mX} \tilde \mX - \mI) \vtheta^\ast 
+
(\mSigma^2 \transp{\tilde \mX} \tilde \mX + \lambda \mI)^{-1} \mSigma^2 \transp{\tilde \mX}\vz,
\\
\mSigma \tilde \vtheta_\lambda - \mSigma \vtheta^\ast 
&=
\mSigma ((\mSigma^2 + \lambda \mI)^{-1} \mSigma^2 - \mI) \vtheta^\ast 
+
(\mSigma^2 + \lambda \mI)^{-1} \mSigma^2 \transp{\tilde \mX}\vz.
\end{align*}

The difference between the residuals is
\begin{align*}
\begin{split}
    \mSigma \vtheta_\lambda - \mSigma \tilde \vtheta_\lambda
    &=
    \mSigma^2 ((\mSigma^2 \transp{\tilde \mX} \tilde \mX + \lambda \mI)^{-1} \transp{\tilde \mX} \tilde \mX - (\mSigma^2 + \lambda \mI)^{-1}) \mSigma \vtheta^\ast 
    \\
    &\qquad \qquad \qquad +
    \mSigma^2 ((\mSigma^2 \transp{\tilde \mX} \tilde \mX + \lambda \mI)^{-1} - (\mSigma^2 + \lambda \mI)^{-1}) \transp{\tilde \mX}\vz.
\end{split}
\\[2ex]
\begin{split}
    &=
    \mSigma^2 (\mSigma^2 \transp{\tilde \mX} \tilde \mX + \lambda \mI)^{-1} (I - \transp{\tilde \mX} \tilde \mX) \mSigma \vtheta^\ast 
    \\
    &\qquad \qquad \qquad +
    \mSigma^2 ((\mSigma^2 \transp{\tilde \mX} \tilde \mX + \lambda \mI)^{-1} - (\mSigma^2 + \lambda \mI)^{-1}) (\mSigma \vtheta^\ast - \transp{\tilde \mX}\vz).
\end{split}
\end{align*}

Where, we added and subtracted $\mSigma^2 (\mSigma^2 \transp{\tilde \mX} \tilde \mX + \lambda \mI)^{-1} \mSigma \vtheta^\ast $ and re-arranged the terms. We bound the norm of the difference between the residuals $\norm[2]{\mSigma \vtheta_\lambda - \mSigma \tilde \vtheta_\lambda}$ by applying Cauchy-Schwarz inequality to the corresponding terms of the RHS of the equation above.
We have, for the first term,
\begin{align*}
    \norm{\mSigma^2 (\mSigma^2 \transp{\tilde \mX} \tilde \mX + \lambda \mI)^{-1} (I - \transp{\tilde \mX} \tilde \mX) \mSigma \vtheta^\ast }
    &\leq
    \norm{\mSigma^2} \norm{(I - \transp{\tilde \mX} \tilde \mX)}
    \norm{(\mSigma^2 \transp{\tilde \mX} \tilde \mX + \lambda \mI)^{-1}}
    \norm[2]{\mSigma \vtheta^\ast}\\
    &\leq
    \max_i \sigma_i^2 \epsilon \frac{1}{\min_i \sigma_i^2 (1-\epsilon) + \lambda} \norm[2]{\mSigma \vtheta^\ast}\\
    &\mystackrel{(i)}{\le}
    2 \epsilon \frac{\max_i \sigma_i^2}{\min_i \sigma_i^2 + \lambda} \norm[2]{\mSigma \vtheta^\ast}
\end{align*}
%
%
where we used
$1 - \epsilon \leq \lVert\transp{\tilde \mX} \tilde \mX\rVert \leq 1 + \epsilon$
and (i) follows by the assumption $\epsilon \leq min_i (\sigma_i^2 + \lambda)/2$
both of which follow from the conditions of the lemma.

We next bound the norm of the second term in the difference between the residuals. We have,

\begingroup
\allowdisplaybreaks
\begin{align*}
    \begin{split}
    &\norm{\mSigma^2 ((\mSigma^2 \transp{\tilde \mX} \tilde \mX + \lambda \mI)^{-1} - (\mSigma^2 + \lambda \mI)^{-1}) (\mSigma \vtheta^\ast - \transp{\tilde \mX}\vz)}\\[2ex]
    &\qquad \qquad \qquad \qquad \leq
    \norm{\mSigma^2}
    \norm{(\mSigma^2 \transp{\tilde \mX} \tilde \mX + \lambda \mI)^{-1} - (\mSigma^2 + \lambda I)^{-1}}
    \norm[2]{\mSigma \vtheta^\ast - \transp{\tilde \mX} \vz}
    \end{split}\\[2ex]
    &\qquad \qquad \qquad \qquad \mystackrel{(i)}{\le}
    \max_i \sigma_i^2
    \norm{(\mSigma^2 \transp{\tilde \mX} \tilde \mX + \lambda \mI)^{-1}}
    \norm{\mSigma^2 (\mI - \transp{\tilde \mX} \tilde \mX)}
    \norm{(\mSigma^2 + \lambda \mI)^{-1}}
    \norm[2]{\mSigma \vtheta^\ast - \transp{\tilde \mX} \vz}\\[2ex]
    &\qquad \qquad \qquad \qquad \leq
    \max_i \sigma_i^2
    \frac{1}{\min_i (\sigma_i^2 (1-\epsilon) + \lambda)}
    \frac{1}{\min_i (\sigma_i^2 + \lambda)}
    \norm{\mSigma^2} \norm{\mI - \transp{\tilde \mX} \tilde \mX}
    \norm[2]{\mSigma \vtheta^\ast - \transp{\tilde \mX} \vz}\\[2ex]
    &\qquad \qquad \qquad \qquad \leq
    2 \epsilon \frac{\max_i \sigma_i^4}{\min_i (\sigma_i^2 + \lambda)^2}
    \left(
        \norm[2]{\mSigma \vtheta^\ast} + \lVert \transp{\tilde \mX} \vz \rVert_2
    \right)
    %
    %
    %
\end{align*}
\endgroup

where the last inequality follows by the assumption $\epsilon \leq min_i (\sigma_i^2 + \lambda)/2$, 
and (i) follows by noting that the matrix $\mSigma^2 \transp{\tilde \mX} \tilde \mX + \lambda \mI$ can be viewed as a perturbation of the non-singular matrix $\mSigma^2 + \lambda \mI$ such that
$\mSigma^2 \transp{\tilde \mX} \tilde \mX + \lambda \mI
=
(\mSigma^2 + \lambda \mI)
-
\mSigma^2 (\mI - \transp{\tilde \mX} \tilde \mX),
$
and applying a standard bound from the literature (see~\cite[Chapter 5, Equation 5.8.1]{horn_MatrixAnalysis_2012}) on the difference of the inverse of the two matrices.
Combining the two bounds yields~\eqref{eq:boundlambdas}, which concludes the proof.


\subsection{Proof of Lemma~\ref{lem:normTildeXTz}}
We have
\begin{align*}
    \norm[2]{\transp{\tilde \mX} \vz}
    =
    \left\lvert{\sum_{l=1}^{d} (\transp{\tilde \vx_l} \vz)^2} \right\rvert^{1/2}
    \leq
    \sum_{l=1}^{d} \norm[2]{\transp{\tilde \vx_l} \vz}
\end{align*}

Conditioned on $\vz$, the random variable $\transp{\tilde \vx}_i \vz$ is zero-mean Gaussian with variance $\norm[2]{\vz}/n$. Thus,
$\PR{ |\transp{\tilde \vx}_i \vz| \geq \frac{\norm[2]{\vz}}{\sqrt{n}}\beta} \leq 2 e^{-\beta^2/2}$.
Moreover, as provided in~\eqref{eq:event2}, with probability at least $1-2e^{-n/8}$, $\norm[2]{\vz}^2 \leq 2 \sigma^2$. 
Combining the two with the union bound, we obtain
\begin{align*}
    \PR{ |\transp{\tilde \vx}_i \vz|^2 \geq \frac{2\sigma^2}{n}\beta^2} \leq 2 e^{-\beta^2/2} + 2 e^{-n/8}.
\end{align*}
Utilizing the union bound again, we obtain
\begin{align*}
    \left\lvert \transp{\tilde \vx_l} \vz \right\rvert
    \leq
    2 \frac{d}{\sqrt{n}} \sigma \beta
\end{align*}
which holds with probability at least $1 - 2d (e^{-\beta^2/2} + e^{-n/8})$.


\subsection{Proof of Lemma~\ref{lem:riskvsexp}}
For proving Lemma~\ref{lem:riskvsexp}, we follow a similar argument to~\cite[Lemma 3]{heckel_EarlyStoppingDeep_2020}.
We have
\begin{align*}
    \risk(\tilde \vtheta_\lambda)
    &=
    \sigma^2 +
    \sum_{i=1}^d \sigma_i^2 
    \underbrace{\left(
        \sigma_i \theta^\ast_i \frac{\lambda}{\sigma_i^2 + \lambda}
        +
        \frac{\sigma_i}{\sigma_i^2 + \lambda} \transp{\tilde \vx_i} \vz
    \right)^2}_{Z_i}.
\end{align*}

Where, $\sum_{i=1}^{d} Z_i$ corresponds to an off-centered chi-squared distribution with the $Z_i$.
The random variable $Z_i$, conditioned on $\vz$, is a squared Gaussian with variance 
upper bounded by $\frac{\norm[2]{\vz}}{\sqrt{n}}$ and has expectation 
\[
\EX{Z_i}
=
\sigma_i^2 (\theta^\ast_i)^2 \left(\frac{\lambda}{\sigma_i^2 + \lambda}\right)^2
+
\frac{\norm[2]{z}^2}{n} \left(\frac{\sigma_i}{\sigma_i^2 + \lambda}\right)^2
\]
By a standard concentration inequality of sub-exponential random variables (see e.g.~\cite[Chapter 2, Equation~2.21]{wainwright_HighDimensionalStatistics_2019}), 
we get, for $\beta \in (0,\sqrt{d})$ and conditioned on $\vz$, that the event
\begin{align}
\label{eq:event1}
\mc E_1
=
\left\{
\left| \sum_{i=1}^d (Z_i - \EX{Z_i}) \right|
\leq
\frac{\norm[2]{\vz}^2}{n} \sqrt{d} \beta
\right\}
\end{align}
occurs with probability at least $1-2e^{ -\frac{\beta^2}{8} }$.
With the same standard concentration inequality for sub-exponential random variables, we have that the event
\begin{align}
\label{eq:event2}
\mc E_2 = 
\left\{
\left| \norm[2]{\vz}^2 - \sigma^2 \right| \leq \frac{\sigma^2 \beta}{\sqrt{n}}
\right\}
\end{align}
also occurs with probability at least $1-2e^{ -\frac{\beta^2}{8} }$. 
By the union bound, both events hold simultaneously with probability at least $1-4e^{ -\frac{\beta^2}{8} }$. 
On both events, we have that
\begin{align*}
    \left| 
        \risk(\tilde \vtheta^\iter) - \bar R(\tilde \vtheta^\iter)
    \right|
    &=
    \left|
        \sum_{i=1}^d (Z_i - \EX{Z_i})
        +
        \frac{1}{n} \left( \norm[2]{\vz}^2 - \sigma^2 \sigma_i^2 \right) \left(\frac{\sigma_i}{\sigma_i + \lambda}\right)^2
    \right| \\
    &\leq
    \left|
        \sum_{i=1}^d (Z_i - \EX{Z_i})
    \right|
    +
    \frac{d}{n}
    \max_i \left[
        \left(\frac{\sigma_i}{\sigma_i + \lambda}\right)^2
        \lvert \norm[2]{\vz}^2 - \sigma^2 \sigma_i^2 \rvert
    \right]
    \\
    &\leq
    \frac{\norm[2]{\vz}^2}{n} \sqrt{d} \beta 
    +
    \frac{d}{n} \frac{1}{\sqrt{n}} \sigma^2 \beta 
    \max_i \left[
        \left(\frac{\sigma_i}{\sigma_i + \lambda}\right)^2 \sigma_i ^ 2
    \right]
    \\
    &\leq
    \frac{2 \sigma^2}{n} \sqrt{d} \beta 
    +
    \frac{d}{n} \frac{1}{\sqrt{n}} \sigma^2 \beta 
    \max_i \left[
        \left(\frac{\sigma_i}{\sigma_i + \lambda}\right)^2 \sigma_i ^ 2
    \right]
    \\
    &\leq
    \frac{2 \sigma^2}{n} \sqrt{d} \beta 
    +
    \frac{d}{n} \frac{1}{\sqrt{n}} \sigma^2 \beta 
    \max_i \left(
        \frac{\sigma_i^2}{\sigma_i + \lambda}
    \right)^2
    \\
    &\mystackrel{(i)}{\le}
    \frac{\sigma^2}{n} \beta
    3 \sqrt{d}.
\end{align*}
where (i) follows from the assumption $d/n \leq \max_i ((\sigma_i + \lambda) / \sigma_i^2)^4$,
which concludes the proof of our lemma.

\subsection{Proof of Proposition 1}
\label{app:proof_prop1}
Here, we provide the formal proof for Proposition 1.

Note that we consider the generalized ridge regression problem, but with a diagonal regularization matrix $\mLambda$ (i.e. Tikhonov regularization).
Specifically, $\mLambda$ is the $\reals^{d \times d}$ diagonal matrix containing regularization parameters $\sqrt{\lambda_i}$ pertaining to each different features along its diagonal.

It then directly follows from the proof of Theorem~1 in Section~\ref{app:proof_theorem_1}, by simply replacing $\lambda \mI$ with ${\mLambda}^{1/2}$, that the risk for the above generalized ridge regression problem is well estimated by the following expression:

\begin{align}
    \label{eq:tikhonov_defbarR}
    \bar \risk(\tilde \vtheta_{\mLambda})
    =
    \sigma^2 + \sum_{i=1}^d
    \underbrace{
    \sigma_i^2 \theta_{i,\ast}^2 
    \left( \frac{\lambda_i}{\sigma_i^2 + \lambda_i} \right)^2  
    +
    \frac{\sigma^2}{n} 
    \sigma_i^2
    \left( \frac{\sigma_i}{\sigma_i^2 + \lambda_i} \right)^2
    }_{V_i({\mLambda})},
\end{align}

We consider the set of values $\{ \lambda_1,\ldots,\lambda_d \}$ that minimizes the risk expression in~\eqref{eq:tikhonov_defbarR}. Since $\bar \risk(\tilde \vtheta_{\mLambda})$ contains a summation of terms pertaining to each feature, we take the derivative of $\bar \risk(\tilde \vtheta_{\mLambda})$ with respect to $\lambda_i$:

\begin{align*}
    \frac{\partial}{\partial \lambda_i} \bar \risk(\tilde \vtheta_{\mLambda})
    &=
    \frac{\partial}{\partial \lambda_i} \left( \sigma^2 + \sum_{j=1}^d V_j({\mLambda}) \right)
    \\
    &=
    \frac{\partial V_i({\mLambda})}{\partial \lambda_i}
    \\
    &=
    2 \sigma_i^2 \theta_{i,\ast}^2 
    \left( \frac{\lambda_i}{\sigma_i^2 + \lambda_i} \right) 
    \frac{(\sigma_i^2 + \lambda_i) - \lambda_i}{(\sigma_i^2 + \lambda_i)^2} 
    -
    2 \frac{\sigma^2}{n} \sigma_i^2 
    \left( \frac{\sigma_i}{\sigma_i^2 + \lambda_i} \right) 
    \frac{\sigma_i}{(\sigma_i^2 + \lambda_i)^2}
    \\
    &=
    \frac{2 \sigma_i^4 \theta_{i,\ast}^2 \lambda_i - 2 \sigma^2 \sigma_i^4 /n}{(\sigma_i^2 + \lambda_i)^3}.
\end{align*}

Setting it above to $0$, we get

\begin{align}
    \label{eq:optimalscalelambda}
    \lambda_i = \frac{\sigma^2}{n} \theta_{i,\ast}^{-2}.
\end{align}

Plugging this back into the expression at~\eqref{eq:tikhonov_defbarR}, we get the risk at the optimal scaling as

\begin{align*}
    \bar \risk(\tilde \vtheta_{\mLambda_{opt}})
    &=
    \sigma^2 + \sum_{i=1}^d
    \sigma_i^2 \theta_{i,\ast}^2 \frac{\sigma^4}{n^2} \theta_{i,\ast}^{-4}
    \left( \frac{1}{\sigma_i^2 + \frac{\sigma^2}{n} \theta_{i,\ast}^{-2}} \right)^2
    +
    \frac{\sigma^2}{n} \sigma_i^2 \left( \frac{\sigma_i}{\sigma_i^2 + \frac{\sigma^2}{n} \theta_{i,\ast}^{-2}} \right)^2
    \\
    &=
    \sigma^2 + \sum_{i=1}^d
    \frac{\sigma^2}{n} \sigma_i^2 \left( \frac{\sigma_i}{\sigma_i^2 + \frac{\sigma^2}{n} \theta_{i,\ast}^{-2}} \right)^2 (\frac{\sigma^2}{n} \theta_{i,\ast}^{-2} + \sigma_i^2)
    \\
    &=
    \sigma^2 + \frac{\sigma^2}{n} \sum_{i=1}^d
    \frac{\sigma_i^2}{\sigma_i^2 + \frac{\sigma^2}{n} \theta_{i,\ast}^{-2}}
    .
\end{align*}

\subsection{Proof of Proposition 2}
Proof of Proposition 2 follows directly by equating the terms in the summation of the risk expression given in (8) for the generalized ridge regression problem and the risk expression of the early-stopped least squares given in (5), as studied in~\citet{heckel_EarlyStoppingDeep_2020}.

It is straightforward to see that the terms inside the respective summations become equal when $\lambda_i$ are chosen as $\lambda_i = \frac{\sigma_i^2}{1 - (1 - \eta_i \sigma_i^2)^{\iter}} -\sigma_i^2$.

\section{Details of how double descent occurs outside the linear regime in neural networks}
\label{app:noNTK}

In this section, we discuss in more detail how the individual parameters of a network with $p$ many parameters trained by applying gradient descent with stepsize $\eta$ to the $\ell_2$-regularized least-squares loss with regularization strength $\lambda$ change across gradient descent iterations.

Note that for an overparameterized network, the network Jacobian $\mJ \in \reals^{n\times p}$ is a wide matrix that typically has full row rank (albeit the small singular values can be very small). 
Let $\mJ = \mU \mSigma \transp{\mV}$ be the singular value decomposition of the Jacobian, where $\mV \in \reals^{p \times n}$ are the right-singular vectors. Note that only the directions of the parameter vector $\vtheta \in \reals^p$ that align with the right-singular vectors $\mV$ impact the predictions of the linear model of the network, however the parameter vector also changes in the directions of the orthogonal complement of the right singular vectors, denoted by $\mV_\perp \in \reals^{p \times (p-n)}$, due to the $\ell_2$-penalty. Specifically, with $\transp{\tilde \mV} = [ \transp{\mV}, \transp \mV_\perp ]$, the parameter update $\vtheta_\iter$ at gradient iteration $\iter$ takes the form
\begin{align*}
    \vtheta_\iter
    &=
    \tilde \mV
    \left(
        \mI - \eta
        \begin{bmatrix}
            \mSigma^2 + \lambda \mI & 0 \\
            0 & \lambda \mI
        \end{bmatrix}
    \right)^\iter
    \transp{\tilde \mV} \vtheta_0
    +
    \eta \sum_{\tau = 0}^{\iter - 1} \tilde \mV
    \left(
        \begin{bmatrix}
            \mSigma^2 + \lambda \mI & 0 \\
            0 & \lambda \mI
        \end{bmatrix}
    \right)^\tau
    \transp{\tilde \mV} \transp \mJ \vy
    \\
    &=
    \tilde \mV
    \left(
        \mI - \eta
        \begin{bmatrix}
            \mSigma^2 + \lambda \mI & 0 \\
            0 & \lambda \mI
        \end{bmatrix}
    \right)^\iter
    \transp{\tilde \mV} \vtheta_0
    +
    \mV
    \text{diag} (
        \ldots,
        \frac{\sigma_i}{\sigma_i^2 + \lambda}
        (1 - (1 - \eta (\sigma_i^2 + \lambda))^t),
        \ldots
    )
    \transp \mU \vy
\end{align*}
Then, the norm of the change in the parameters that is relevant to fitting the data is
\begin{align}
    \norm[2]{\transp \mV (\vtheta_\iter - \vtheta_0)}^2
    =
    \sum_i^n 
    (1 - (1 - \eta (\sigma_i^2 + \lambda))^\iter)^2
    \left(
        -\frac{1}{\sigma_i} \innerprod{\vu_i}{\mJ \theta_0}
        +
        \frac{\sigma_i}{\sigma_i^2 + \lambda} \innerprod{\vu_i}{\vy}
    \right)^2.
\end{align}
Note that the convergence rate for the above depends primarily on the smallest singular value $\sigma_{\min}$.
For a sufficiently small stepsize, we have $(1 - \eta (\sigma_i^2 + \lambda))^\iter \approx \exp(-\eta \iter (\sigma_i^2 + \lambda))$, which means that this part converges when $\exp(-\eta \iter (\sigma_{\min}^2 + \lambda))$ gets close to zero. This is the part that is relevant to fitting the data and if initialized appropriately, this change is not more than $O(n)$. 

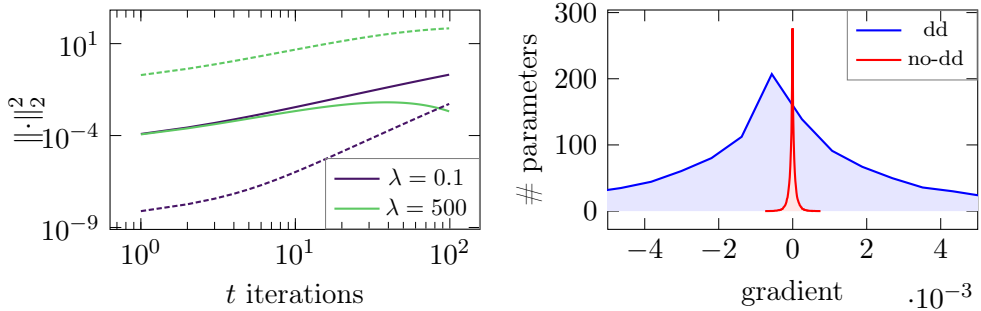
\begin{figure}[t]
\begin{center}
\hspace{-0.8cm}
\begin{tikzpicture}

\begin{groupplot}[
width=6.5cm,
height=4.5cm,
group style={
    group name=resnet,
    group size= 2 by 1,
    xlabels at=edge bottom,
    horizontal sep=1.7cm, 
    vertical sep=1.1cm,
},
xlabel={gradient},
ylabel style={at={(-0.15,0.5)}},
xtick style={color=black},
y grid style={white!69.01960784313725!black}, 
ytick style={color=black},
legend columns=1,
legend style={
            at={(1,1)}, 
            anchor=north east, 
            draw=black!50, 
            fill=none,
            font=\scriptsize,
            /tikz/every even column/.append style={column sep=0.3cm}
        },
colormap name=viridis,
cycle list={[colors of colormap={50,250,450,700}]},
]

\nextgroupplot[xlabel={$\iter$ iterations},
            ylabel={$\norm[2]{\cdot}^2$},
            legend style={
                nodes={scale=0.9, transform shape},
                at={(1,0)}, 
                anchor=south east,
                draw=black!50,
                fill=none,
                font=\small
            },
            xmode=log,
            ymode=log,
        ]
    
    \addplot +[mark=none, thick, color of colormap={50}] table[x=t,y expr=\thisrow{wd0-1-vtheta},select coords between index={0}{100}]{./fig/data/vtheta_paramater_change.txt};
    \addlegendentry{$\lambda=0.1$}

    \addplot +[mark=none, thick, color of colormap={750}] table[x=t,y expr=\thisrow{wd500-vtheta},select coords between index={0}{100}]{./fig/data/vtheta_paramater_change.txt};
    \addlegendentry{$\lambda=500$}


    \addplot +[mark=none, thick, color of colormap={50}, dash pattern=on 2pt off 1pt] table[x=t,y expr=\thisrow{wd0-1-tildevtheta},select coords between index={0}{100}]{./fig/data/vtheta_paramater_change.txt};

    \addplot +[mark=none, thick, color of colormap={750}, dash pattern=on 2pt off 1pt] table[x=t,y expr=\thisrow{wd500-tildevtheta},select coords between index={0}{100}]{./fig/data/vtheta_paramater_change.txt};

\nextgroupplot[
    ylabel={\# parameters},
    xmin=-0.005, xmax=0.005,
]


\addplot [name path=p0t0no, thick, blue, mark color=blue]
table[x=p0t0bins, y=p0t0hist] {fig/data/dd_gradients.txt};
\addlegendentry{dd}

\addplot [name path=p0t0dd, thick, red, mark color=red]
table[x=p0t0bins, y expr=\thisrow{p0t0hist}/40] {fig/data/no-dd_gradients.txt};
\addlegendentry{no-dd}


\addplot [name path=low0dd,draw=none] table[x=p0t0bins, y expr=0] {fig/data/dd_gradients.txt};
\addplot [fill=blue!10] fill between[of=p0t0no and low0dd];

\addplot [name path=low0no,draw=none] table[x=p0t0bins, y expr=0] {fig/data/no-dd_gradients.txt};
\addplot [fill=red!10] fill between[of=p0t0dd and low0no];


\end{groupplot}

\end{tikzpicture}%
\end{center}
\caption{
\label{fig:twolayergradients}
{\bf Left:}
The norm of the change in the parameters that is relevant to fitting the data ({\it solid}) and not relevant to fitting the data ({\it dashed}) for large and small values of $\lambda$. The results show that the parameters primarily change in the directions that are not relevant for fitting the data when $\lambda$ becomes larger. This moves the neural network outside of the NTK regime (see SM~\ref{app:noNTK} for details). 
{\bf Right:}
Distribution of the gradients corresponding to the first layer parameters of the network at the first gradient iteration ($t=1$) for $\lambda=0.001$.
The \textcolor{red}{red curve} (scaled back $\sim$3 times for the sake of visualization) corresponds to the data setup where the difference in the scales of the data features is suppressed, hence resulting in no double descent behavior. The \textcolor{blue}{blue curve} corresponds to the setting where the features are scaled as discussed before with double descent present as a function of the regularization strength. The results indicate that the dynamics of the network is different from the very beginning for the two regimes even for small $\lambda$.
}
\vspace{-0.35cm}
\end{figure}

We next consider the change of the coefficient vector that is not relevant to fitting the training data:
\begin{align}
    \norm[2]{\transp \mV_\perp (\vtheta_\iter - \vtheta_0)}^2
    &=
    (1 - (1 - \eta \lambda)^\iter)^2 \norm[2]{\transp \mV_\perp \vtheta_0}^2
    \\
    &\approx
    (1 - e^{-\eta \lambda \iter})^2 O(p). \nonumber
\end{align}
Therefore, the change in the coefficients for any $\lambda$ is on the order of p, and hence is not contained within a small radius around the initialization, where the NTK approximation accurately captures the dynamics of the nonlinear network, unless $1 - e^{-\eta \lambda \iter}$ is very small (see Figure~\ref{fig:twolayergradients} (left) for an illustration).

In order to observe how this translates to the relationship between the smallest singular value of the network Jacobian $\sigma_\text{min}$, and $\lambda$, consider the following assumption on $1 - e^{-\eta \lambda \iter}$ being sufficiently small as parameterized by a small number $\delta$, i.e. $1 - e^{-\eta \lambda \iter} \leq \delta$. We then have $\lambda \leq \frac{-1}{\eta \iter} \ln (1-\delta) \approx \frac{\delta}{\eta \iter}$.
Note that we are also interested in the training regime until the network is close to convergence. This occurs when $\exp(-\eta \iter (\sigma_{\min}^2 + \lambda)) \approx 0$ or $\exp(-\eta \iter (\sigma_{\min}^2 + \lambda)) \leq \epsilon$ for small $\epsilon$. This in turn leads to the condition $\sigma_{\min}^2 \geq \frac{1/\epsilon - \delta}{\eta \iter}$.

Based on these conditions on the $\sigma_{\min}$ and $\lambda$, in order for the change in the parameters to be confined in a small radius around the network initialization, we need $\sigma_{\min}^2 \gg \lambda$.
Based on our empirical observations, in the regime where double descent is observed, $\lambda$ is much greater than $\sigma_{\min}^2$ and the above condition does not hold.

While in this section we study how the parameters of a network change throughout the training for any $\lambda$ with respect to a fixed kernel, a similar result was shown for how the associated neural tangent kernel changes across gradient flow time $\iter$ (iterations) with respect to $\lambda$ (see~\cite[Theorem 1]{lewkowycz2021TrainingDynamicsDeep}).
Specifically, \citet{lewkowycz2021TrainingDynamicsDeep} have shown that, when gradient flow is applied to the $\ell_2$-regularized MSE loss, the singular values of the kernel decay exponentially from the initialization with respect to $\lambda \iter$, whereas the singular vectors remain static. 
This is in agreement with our discussion that $\sigma_{\min}^2 \gg \lambda$ is needed for a fixed kernel at initialization to accurately capture the training dynamics of the non-linear network throughout the course of the gradient descent.


Lastly, we show that even for small $\lambda$, the linearization (or NTK approximation) is not a good approximation for 
the network in a setup where regularization-wise double descent occurs. 
Specifically, when the disparity between the variances across the features of the data is sufficiently large to yield double descent, the change in the parameters of the network is large even for small $\lambda$.
This can be seen in Figure~\ref{fig:twolayergradients} (right) for a two layer neural network. As indicated by the blue curve here, in the setting where the underlying data structure has differently scaled features and double descent is observed, the parameters change significantly from the initialization early on during the training even at smaller regularization strength. Note that, based on the decay of the kernel, this is not projected to occur until $t\sim 10^3$ for $\lambda=0.001$ given in this example.

\end{document}